\documentclass{article}
\usepackage{ijcai24}

\usepackage{times}
\usepackage{soul}
\usepackage{url}
\usepackage[hidelinks]{hyperref}
\usepackage[utf8]{inputenc}
\usepackage[small]{caption}
\usepackage{graphicx}
\usepackage{amsmath}
\usepackage{amsthm}
\usepackage{xcolor}
\usepackage{tcolorbox}
\usepackage{booktabs}
\usepackage{algorithm}
\usepackage{algorithmic}
\usepackage[switch]{lineno}
\usepackage{subcaption}
\usepackage{multirow}

\title{Learning to Solve Geometry Problems via Simulating \\Human Dual-Reasoning Process}

\author{
Tong Xiao$^{1}$
\and
Jiayu Liu$^{1}$\and
Zhenya Huang$^{1,2}$\and
Jinze Wu$^4$\and \\
Jing Sha$^4$\and
Shijin Wang$^{3,4}$\and
Enhong Chen\thanks{Corresponding author.}$^{1,3}$
\affiliations
$^1$University of Science and Technology of China \\
$^2$Institute of Artificial Intelligence, Hefei Comprehensive National Science Center \\
$^3$State Key Laboratory of Cognitive Intelligence \\
$^4$iFLYTEK AI Research\\
\emails
\{tongxiao2002, jy251198, hxwjz\}@mail.ustc.edu.cn,
\{huangzhy, cheneh\}@ustc.edu.cn,
\newline
\{jingsha, sjwang3\}@iflytek.com
}

\begin{document}

\maketitle

\begin{abstract}
    Geometry Problem Solving (GPS), which is a classic and challenging math problem, has attracted much attention in recent years. It requires a solver to comprehensively understand both text and diagram, master essential geometry knowledge, and appropriately apply it in reasoning. However, existing works follow a paradigm of neural machine translation and only focus on enhancing the capability of encoders, which neglects the essential characteristics of human geometry reasoning. In this paper, inspired by dual-process theory, we propose a \textbf{Dual}-Reasoning \textbf{Geo}metry \textbf{Solver} \textbf{(DualGeoSolver)} to simulate the dual-reasoning process of humans for GPS. Specifically, we construct two systems in DualGeoSolver, namely \textit{Knowledge System} and \textit{Inference System}. Knowledge System controls an implicit reasoning process, which is responsible for providing diagram information and geometry knowledge according to a step-wise reasoning goal generated by Inference System. Inference System conducts an explicit reasoning process, which specifies the goal in each reasoning step and applies the knowledge to generate program tokens for resolving it. The two systems carry out the above process iteratively, which behaves more in line with human cognition. We conduct extensive experiments on two benchmark datasets, GeoQA and GeoQA+\footnote{The source code and datasets are available at \texttt{\url{https://github.com/tongxiao2002/DualGeoSolver}}}. The results demonstrate the superiority of DualGeoSolver in both solving accuracy and robustness from explicitly modeling human reasoning process and knowledge application.
\end{abstract}

\section{Introduction}
\begin{figure}[t]
    \centering
    \includegraphics[width=\linewidth]{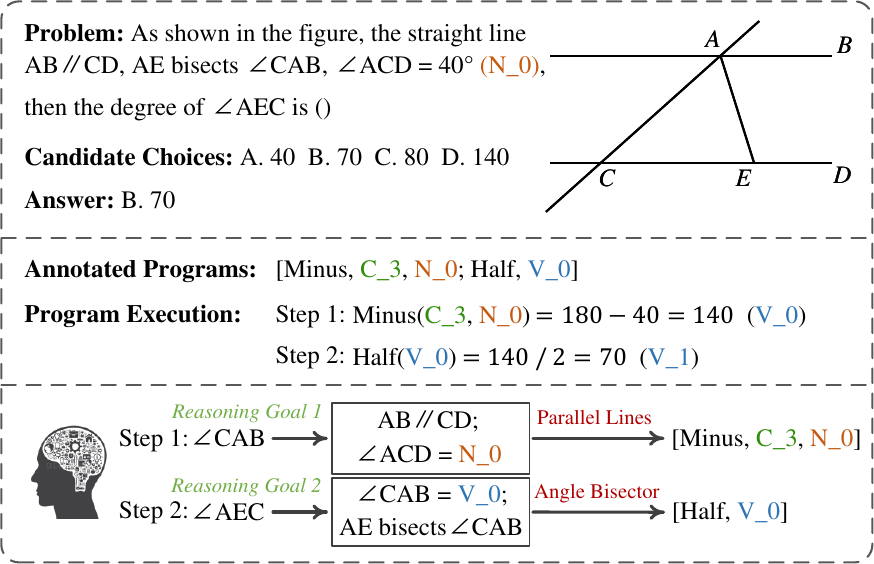}
    \caption{A typical geometry problem in GeoQA dataset.}
    \label{fig:intro-case}
\end{figure}

Automatically solving math problems with AI techniques has attract much attention recently \cite{xie2019goal,zhang2020graph,liang2021solving,liu2022cognitive}, which is considered a crucial step towards achieving general artificial intelligence.
Among various math problems, geometry problem solving (GPS) stands out as a classic and challenging task that demands the ability of multi-modal understanding and reasoning. As shown in the upper part of Figure \ref{fig:intro-case}, a typical geometry problem consists of a textual problem description (``As shown ... $\angle{AEC}$ is ()'') and a geometry diagram (the image of $A,B,C,D,E$), and requires solvers to select an option from four given candidates as the final answer (``B. 70''). This process requires solvers to have a comprehensive understanding of text and diagram, and possess essential geometry knowledge to perform multi-modal reasoning.

Existing works about GPS can be divided into two genres: symbolic geometry solvers and neural geometry solvers. The symbolic solvers \cite{Seo2014DiagramUI,Seo2015SolvingGP,Lu2021InterGPSIG} typically depend on the handcrafted rules to parse the problem text and diagram into logical formal language and then solve the problem by logical deduction. Although having strong interpretability, they encounter severe generalization problems in large scale datasets. Recently, with the rising usage of deep learning for automated math problem solving, many neural geometry solvers \cite{Chen2021GeoQAAG,Cao2022AnAB,Ning2023ASC} have been developed. Neural solvers adopt a encoder-decoder framework to solve geometry problems in a Sequence-to-Sequence way. They encode the problem text and diagram and then feed them into a program decoder to generate a program sequence as shown in ``Annotated Program'' of Figure \ref{fig:intro-case} (i.e. ``[Minus, C\_3, N\_0; Half, V\_0]''), obtaining the final answer through program execution. This line of research has achieved remarkable progress in GPS, with relatively great performance and generalization ability.
However, it still follows the paradigm of neural machine translation which is greatly different from the geometry reasoning process of humans.
For instance, when humans solve the problem in Figure 1, they first recognize that the reasoning goal in Step 1 is to ``calculate $\angle{CAB}$". For this purpose, they note that $AB\ \mathbin{\!/\mkern-4mu/\!}\ CD$ and would recall the geometry knowledge “If the two lines are parallel, then consecutive interior angles are supplementary” from their mind. Guided by this knowledge and realizing $\angle{ACD} = 40^\circ$, they finally deduce that the measure of $\angle{CAB}$ could be obtained by subtracting the measure of $\angle{ACD}$ from $180^\circ$ (i.e., ``C\_3'' in Figure \ref{fig:intro-case}). Existing methods do not explicitly model this reasoning process, and only learn a fitting pattern from problems to programs, while lacking application of the geometry knowledge in reasoning, which may result in model confusion and robustness issues.
Therefore, in this paper, we aim to explicitly model the above human reasoning behaviors to achieve more reliable reasoning process for GPS.

To this end, we draw insights from dual-process theory~\cite{schneider1977controlled,Evans2008DualprocessingAO,kahneman2011thinking,Lieto2017DualPA} of human cognition, which states that there exist two cognitive systems underlying human reasoning, including System 1 and System 2. System 1 represents an implicit process, which is heuristics and retrieves information in a rapid and unconscious way. System 2 represents an explicit process, which is analytic and conducts a slow but controlled thinking process. Combined with the above example, in terms of GPS, System 1 is responsible for providing information about geometry primitives in diagram and appropriate geometry knowledge. System 2 conducts analytic and sequential reasoning by generating step-wise reasoning goals for the problem and integrates the knowledge of System 1 to resolve them.
However, when attempting to simulate this human reasoning process in neural solvers, we may encounter three challenges. Firstly, the reasoning goal serves as the core that guides the whole reasoning process. It is crucial but difficult to precisely identify and update the reasoning goal. Secondly, the geometry knowledge used may vary throughout the entire reasoning process, making it challenging to correctly select geometry knowledge for different reasoning steps. Thirdly, after obtaining the geometry knowledge, it is also a challenge to integrate it with reasoning goal and apply them to guide the explicit reasoning in System 2.

To tackle the aforementioned issues, we propose a \textbf{Dual}-Reasoning \textbf{Geo}metry \textbf{Solver} \textbf{(DualGeoSolver)}, which explicitly models human dual-reasoning process in GPS. Specifically, we construct two systems in DualGeoSolver, namely \textit{Knowledge System} and \textit{Inference System}, which simulate System 1 and System 2, respectively. At the beginning of each reasoning step, Knowledge System provides diagram information and geometry knowledge according to the current reasoning goal. For the former, we propose a Visual Spotlight Module that captures the relationships between reasoning goal and geometry primitives in the diagram, while for the latter we propose another Knowledge Selection Module to retrieve sufficient and relevant knowledge from an external knowledge base. Furthermore, we design a Knowledge Injection Module to aggregate diagram information and geometry knowledge to direct the reasoning process in Inference System.
In Inference System, we first apply the the knowledge from Knowledge System to generate target program tokens for resolving the current reasoning goal. Then, we develop a novel Goal Generation Module to identify the next solvable reasoning goal based on known conditions and preceding reasoning information as humans.
This goal is then fed back into Knowledge System to start the next reasoning step. By alternately repeating the above two processes, our DualGeoSolver eventually solves the original geometry problem in a human-like manner.

In summary, the contributions of this paper include:
\begin{itemize}
    \item We propose a novel DualGeoSolver that constructs Knowledge-Inference systems to model human dual-reasoning process in solving geometry problems.
    \item We propose elaborate knowledge selection and injection mechanisms to simulate the implicit reasoning process of humans, and design a novel goal-oriented manner to achieve explicit reasoning process.
    \item Extensive experiments on two public datasets GeoQA and GeoQA+ demonstrate the superiority of our DualGeoSolver compared with 7 GPS baselines and 7 LLMs.
\end{itemize}

\section{Related Works}
Existing works on GPS could be divided into two genres: Symbolic Geometry Solvers and Neural Geometry Solvers.

\subsection{Symbolic Geometry Solvers}
\cite{Seo2015SolvingGP} proposed the first symbolic solver GeoS, which first parsed the problem text and diagram into first-order logic literals using handcrafted rules and OCR techniques \cite{Seo2014DiagramUI}, then solved the geometry problem by finding an assignment that satisfied all the parsed literals. In order to alleviate the reliance of GeoS on handcrafted rules, \cite{Sachan2017FromTT,Sachan2017LearningTS} injected geometry theorem knowledge as the form of horn-clauses into the solver and replaced the handcrafted rules. Recently, Inter-GPS \cite{Lu2021InterGPSIG} improved the reasoning process of previous symbolic solves by iteratively searching geometry primitives and applying a series of manually defined geometry theorems. Although symbolic solvers have achieved significant progress and possess strong interpretability, they heavily rely on the handcrafted rules to parse the geometry problems and lack generalization.

\subsection{Neural Geometry Solvers}
With the rising usage of deep learning for automated math problem solving, \cite{Chen2021GeoQAAG} first proposed a neural geometry solver called NGS which solved the geometry problems with an encoder-decoder framework. It encoded the problem text and diagram separately, then fused them using a multi-modal fusion module, and finally sent them to a program decoder to generate program tokens that can produce final numeric result through program execution. On this basis, to improve the text encoder, DPE-NGS \cite{Cao2022AnAB} adopted both Bi-LSTM and RoBERTa\cite{Liu2019RoBERTaAR} for encoding, which were further enhanced by SCA-GPS \cite{Ning2023ASC} through integrating diagram features with symbolic characters. Geoformer \cite{Chen2022UniGeoUG} adopted T5 \cite{Raffel2019ExploringTL} model as the backbone and strengthened the reasoning ability by introducing geometry proving problems. For precisely describing the diagram, \cite{Zhang2022PlaneGD} proposed PGDPNet which utilized instance segmentation \cite{he2017mask,ying2021embed} and scene graph generation \cite{xu2017scene} techniques to parse the geometry primitives and their relations from the diagram. Subsequently, to better understand the semantics meanings of different geometry primitives, PGPSNet \cite{Zhang2023AMN} solved geometry problems by applying semantic embeddings to different types of geometry primitives. Though neural solvers have achieved remarkable performance and possessed strong generalization ability, they still solve geometry problems by following a neural machine translation paradigm, while neglecting the characteristics of human reasoning in geometry problem solving, which may lead to model confusion and robustness issues. 
Differently, in this paper, we draw insights from dual-process theory and simulate human dual-reasoning process to solve geometry problems.


\section{Methodology}
\begin{figure*}[t]
    \centering
    \includegraphics[width=\linewidth]{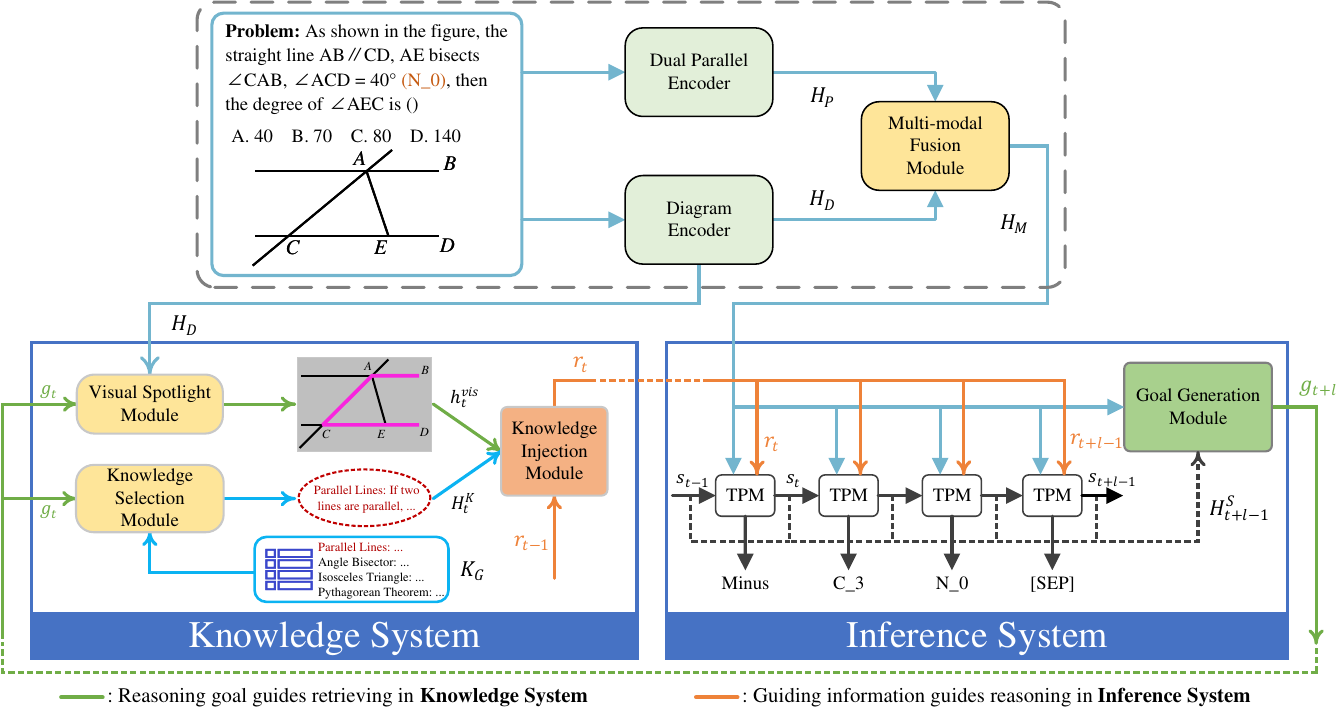}
    \caption{Overview of our DualGeoSolver. The whole dual-reasoner consists of a Knowledge System and an Inference System. We illustrate the first reasoning step with the cooperation of two systems, which finally generates program tokens ``[Minus, C\_3, N\_0]''.}
    \label{fig:framework}
\end{figure*}

\subsection{Problem Definition}
Formally, a geometry problem is defined as a tuple ($D$, $P$, $\boldsymbol{c}$), where $D$ represents the geometry diagram, $P = [p_1, p_2, \dots, p_n]$ represents $n$ tokens in the problem description, and $\boldsymbol{c} = \{c_1, c_2, c_3, c_4\}$ represents the multi-choice candidates where each choice is a numeric value. Given the geometry diagram $D$ and problem description $P$, one GPS solver is trained to select a choice $c_i \in \boldsymbol{c}$.

In the current field of GPS, researchers do not make neural solvers predict a choice from candidates $\boldsymbol{c}$ directly.
Alternately, they annotate a program as a sequence of tokens $Y_P = [y_1, y_2, \dots, y_T]$ that can be executed to obtain a numeric result. Each $y_t\in Y_P$ comes from a program vocabulary $V_P$ composed of four parts: the operators $V_O$ where each operator represents a mathematical operation (e.g., ``Minus"), the numeric constants $V_C$ (e.g., ``C\_3" which stands for $180^\circ$), the numeric values $N_P$ which appear in problem description $P$ (e.g. ``N\_0" which stands for $40^\circ$), and the numeric variables $N_V$ which are the intermediate execution results of previous reasoning steps (e.g., ``V\_0" which stands for $140^\circ$). That is, $V_P = V_O \cup V_C \cup N_P \cup N_V$.
It is worth noting that $N_P$ and $N_V$ are constructed by number mapping, which transforms the numerical values into a unified representation.

\textit{Definition 3.1.} Given a geometry problem ($D$, $P$, $\boldsymbol{c}$), our goal is to build a model that could reason the program $Y_P = [y_1, y_2, \dots, y_T]$, then obtain the numeric result $z$ through executing the program $Y_P$, and finally select a choice $c_i$ matching $z$ from candidates $\boldsymbol{c}$.

\subsection{Overall Framework}
To explicitly model human reasoning process, inspired by the dual-process theory, we propose a \textbf{Dual}-Reasoning \textbf{Geo}metry \textbf{Solver} \textbf{(DualGeoSolver)} as depicted in Figure \ref{fig:framework}.
Firstly, we encode the problem text $P$ and diagram $D$ through separate encoders and fuse them through a multi-modal fusion module. Then, we feed them into our dual-reasoner to iteratively generate the target program $Y_P = [y_1, y_2, \dots, y_T]$ with the cooperation of Knowledge-Inference systems, and obtain the numeric result through program execution.

\subsection{Problem Encoders}
We adopt two encoders to extract the features of diagram $D$ and problem text $P$ separately, and then fuse and align them.

\textbf{Diagram Encoder.}
To extract visual information from geometry diagram $D$,
We employ a ViTMAE \cite{He2022MaskedAA} that handles geometry diagram through two pre-training tasks, including Masked Image Modeling (MIM) and symbolic character detection \cite{Ning2023ASC}.
Given diagram $D$, we divide it into regularly non-overlapping $\gamma \times \gamma$ patches and input them into ViTMAE. We take the outputs of last hidden layer as the visual features $H_D = [h_1^D, h_2^D, \dots, h_m^D]$, where $m = \gamma \times \gamma$ is the number of diagram patches.

\textbf{Text Encoder.} Given textual problem description with $n$ tokens $P = [p_1, p_2, \dots, p_n]$, we feed it into a LSTM \cite{Hochreiter1997LongSM} and RoBERTa \cite{Liu2019RoBERTaAR} separately to obtain richer representations of $P$. Then, we aggregate them through a linear layer to obtain the final encoded textual features $H_P = [h_1^P, h_2^P, \dots, h_n^P]$.

\textbf{Multi-modal Fusion Module.}
We employ a multi-modal co-attention module \cite{Yu2019DeepMC} to fully fuse and align the diagram features $H_D$ and text features $H_P$.
The co-attention module takes $H_D$ and $H_P$ as inputs and outputs the multi-modal representation $F_D = [f_1^D, f_2^D, \dots, f_m^D]$. We concatenate $H_P$ and $F_D$ to form $H_M = [H_P; F_D] = [h^M_1, h^M_2, \dots, h^M_{n + m}]$ for subsequent reasoning.
Additionally, we apply an attention-reduction network \cite{Chen2021GeoQAAG} to aggregate $F_D$ into a vector. We then concatenate this vector with the features of the last token in $H_P$, obtaining the aggregated multi-modal feature vector $h_M$.

\subsection{Dual-Reasoner}
Existing geometry solvers decode the program $Y_P$ in a totally Seq2Seq manner, which is greatly different from human reasoning process. Taking Figure \ref{fig:intro-case} as an example, humans start by identifying the first reasoning goal as ``calculate $\angle{CAB}$''. Then, they conduct implicit reasoning from two aspects: 1) refer to the diagram and capture relationships between $\angle{CAB}$ and other geometry primitives, 2) retrieve the geometry knowledge such as ``Parallel Lines". Under the guidance of them, humans conduct explicit reasoning and deduce tokens ``[Minus, C\_3, N\_0]'' to solve this goal. Afterwards, they realize the next goal is to ``calculate $\angle{AEC}$' and repeat the above processes until obtaining the final answer.

To achieve the above processes, we draw insights from dual-process theory \cite{schneider1977controlled,Evans2008DualprocessingAO,kahneman2011thinking,Lieto2017DualPA} to propose a dual-reasoner which conduct a dual-reasoning process with two systems, namely \textit{Knowledge System} and \textit{Inference System}. Specifically, Knowledge System represents the implicit reasoning process. At time step $t$, it receives reasoning goal $g_t$ generated by Inference System, and retrieves both geometry knowledge and diagram information to form an overall guiding information $r_t$. Inference System represents the explicit reasoning process. It resolves the reasoning goal $g_t$ based on $r_t$, which sequentially generates $l$ program tokens that can solve $g_t$ through execution until it reaches a delimiter (e.g. ``$\left[\textnormal{Minus, C\_3, N\_0, [SEP]}\right]$'' in Figure \ref{fig:framework}). Subsequently, Inference System identifies a new reasoning goal $g_{t + l}$ and feeds it into Knowledge System to start next reasoning step. These two systems alternate until completing the entire reasoning process. In the following parts, we introduce them in detail.



\subsubsection{Knowledge System}
In Knowledge System, we design three modules: Knowledge Selection Module (KSM), Visual Spotlight Module (VSM) and Knowledge Injection Module (KIM). At time step $t$, KSM receives reasoning goal $g_t$ from Inference System and then retrieves geometry knowledge $H^{K}_t$ that contributes to resolving the reasoning goal from an external knowledge base.
Meanwhile, VSM retrieves diagram information $h^{vis}_t$ from the diagram based on $g_t$, which captures the relationships between reasoning goal and geometry primitives in the diagram. Finally, KIM integrates $H^{K}_t$, $h^{vis}_t$ and previous reasoning state (denoted as $s_{t - 1}$) from Inference System to form a guiding information $r_t$ that directs the subsequent explicit reasoning process in Inference System.

\textbf{Knowledge Selection Module.} Given a reasoning goal $g_t$ generated by Inference System, we utilize Knowledge Selection Module (KSM) to retrieve relevant geometry knowledge (e.g., ``Parallel Lines'').
To the best of our knowledge, there hasn't been a dedicated geometry knowledge collection.
so we first manually build a geometry knowledge base $K_G$. Each item in $K_G$ is a knowledge-explanation pair, where the ``knowledge'' represents a geometry knowledge concept, and the ``explanation'' provides a textual description of the knowledge. For example, the knowledge concept ``Parallel Lines" in $K_G$ corresponds to the textual description ``If the two lines are parallel, then consecutive interior angles are supplementary".
To build $K_G$ from scratch, we initially gathered the geometry knowledge annotated in GeoQA and GeoQA+ datasets, then collected the detailed explanations of these knowledge concepts from WikiPedia \footnote{\href{https://www.wikipedia.org/}{https://www.wikipedia.org/}} and Baidu-Baike \footnote{\href{https://baike.baidu.com/}{https://baike.baidu.com/}}. Subsequently, these explanations were manually verified for correctness and underwent optimization of expression and logic by three well-trained annotators with undergraduate degrees.

Formally, for each knowledge-explanation pair $<k^p_i, k^e_i>$ in knowledge base $K_G$, we denote its knowledge concept as $k^p_i$ and corresponding explanation as $k^e_i$:
\begin{equation}
    K_G = \bigcup_{i=1}^N \left\{<k^p_i, k^e_i>\right\},\ N=|K_G|
\end{equation}

We model knowledge selection as a multi-label classification task. Therefore, we input $g_t$ into a linear layer with sigmoid activation to obtain the prediction score for each geometry knowledge. We then select and concatenate explanations whose prediction score are higher than a pre-defined threshold $\theta$ for further reasoning. Finally, the selected explanations will be fed into the same text encoder of problem description to obtain knowledge representation $H^{K}_t = \left[u^t_1, u^t_2, \dots, u^t_{n_c}\right]$, where $n_c$ represents the length of concatenated explanations.


\textbf{Visual Spotlight Module.} Given the reasoning goal $g_t$, humans also refer to the diagram to capture essential relationships between geometry primitives in the diagram (e.g. notice that $\angle{ACD}$ and $\angle{CAB}$ are consecutive interior angles in Figure \ref{fig:framework}). We simulate this behavior through Visual Spotlight Module (VSM).
To let Knowledge System pay more attention to the geometry primitives related to the reasoning goal $g_t$ rather than other meaningless places,
we utilize an attention mechanism to fuse $g_t$ and the features of geometry diagram $H_D$, and obtain the visual context vector $h^{vis}_t$.
\begin{equation}
    h^{vis}_t = \sum_{i} h^D_i\dfrac{\exp{\left(g_t \cdot h^D_i\right)}}{\sum_{j} \exp{\left(g_t \cdot h^D_j\right)}}
\end{equation}

\textbf{Knowledge Injection Module.} Knowledge Injection Module (KIM) aims to integrate the knowledge $H^{K}_t$ and visual context $h^{vis}_t$ to form a guiding information $r_t$. We first utilizes an attention mechanism to aggregate $H^{K}_t$ into vector $h^{know}_t$ based on $r_{t - 1}$, then we employ a single-layer unidirectional LSTM network $\zeta$ to integrates $h^{know}_t$, $h^{vis}_t$ and the previous reasoning state $s_{t - 1}$ from Inference System (described later in Eq.~\eqref{s_t}), updating $r_{t - 1}$ into $r_t$ ($r_0$ is obtained by feeding $h_M$ into a linear layer):
\begin{equation}
    h^{know}_t = \sum_i u^t_i\cdot \dfrac{\exp\left(r_{t - 1}\cdot u^t_i\right)}{\sum_j \exp\left(r_{t - 1}\cdot u^t_j\right)}
\end{equation}

\begin{equation}
    r_t = \zeta\left(\left[h^{vis}_t, h^{know}_t, s_{t - 1}\right], r_{t - 1}\right)
\end{equation}

\subsubsection{Inference System}
In Inference System, we design two modules: Token Prediction Module (TPM) and Goal Generation Module (GGM). TPM combines the multi-modal features $H_M$ and guiding information $r_t$ from Knowledge System to sequentially generate the target program tokens $\{y_i\} (t \leq i < t + l)$ until it reaches a pre-defined delimiter (e.g. ``[SEP]" token in Figure \ref{fig:framework}). Here, $l$ represents the number of program tokens generated by TPM in the current reasoning step. Next, GGM finds the next reasoning goal $g_{t + l}$ according to the known conditions and the preceding reasoning information until time step $t + l - 1$. It then sends $g_{t + l}$ back to Knowledge System to start the next reasoning step. For the sake of symbol consistency, we denote next reasoning goal as $g_{t + l}$ rather than $g_{t + 1}$.

\textbf{Token Prediction Module.} Token Prediction Module (TPM) is designed to generate target program tokens $\{y_i\} (t \leq i < t + l)$ which could resolve the current reasoning goal $g_t$ through program execution. We employ an LSTM network $\varphi$ with attention mechanism \cite{Bahdanau2014NeuralMT} as our TPM. At time step $t$, $\varphi$ updates its hidden state $s_{t - 1}$ according to the multi-modal features $H_M$, guiding information $r_t$, and the embedding of last generated token $e_{t - 1}$:

\begin{equation}
    h^c_t = \sum_i h^M_i\cdot \dfrac{\exp\left(s_{t - 1}\cdot h^M_i\right)}{\sum_j \exp\left(s_{t - 1}\cdot h^M_j\right)}
\end{equation}

\begin{equation}\label{s_t}
    s_t = \varphi\left(\left[h^c_t, r_t, e_{t - 1}\right], s_{t - 1}\right)
\end{equation}

Then we feed $s_t$ into a linear layer with softmax function to predict the distribution of next program token $P_t$. It is worth noting that the guiding information $r_t$ will not change within a reasoning step, i.e., $r_t = r_{t + 1} = \cdots = r_{t + l - 1}$.

\textbf{Goal Generation Module.} Goal Generation Module (GGM) is designed to identify the next solvable reasoning goal according to the known conditions and preceding reasoning information. Specifically, we consider the multi-modal features $H_M$ as the known conditions since it contains rich information about the raw problem. We regard $H^S_{t + l - 1} = [s_0, s_1, \dots, s_{t + l - 1}]$ containing all the preceding reasoning information. Then we employ a 2-layer Transformer-Decoder \cite{Vaswani2017AttentionIA} network to fully integrate $H_M$ and $H^S_{t + l - 1}$, where we treat $H^S_{t + l - 1}$ as query and $H_M$ as key and value, and take the last vector of the outputs as the next reasoning goal $g_{t + l}$. When $t = 0$, $H^S_0$ degenerates to only include the ``[BOS]'' token, which means GGM only takes known conditions of the question to determine the first solvable reasoning goal.

In summary, the training objectives of DualGeoSolver come from two parts, a generation loss $\mathcal{L}_g$ from Token Prediction Module and a multi-label classification loss $\mathcal{L}_c$ from Knowledge Selection Module. Specifically, $\mathcal{L}_g$ is the negative log-likelihood (NLL) of generating target program tokens:
\begin{equation}
    \mathcal{L}_g = -\frac{1}{T}\sum_{t = 1}^T \log P_t\left(y_t | H_D, H_P, y_1, y_2, \dots, y_{t - 1}\right)
\end{equation}

The multi-label classification loss is the sum of binary cross entropy (BCE) loss of all reasoning steps:
\begin{equation}\label{loss_c}
    \mathcal{L}_c = -\sum_{i = 1}^S\sum_{j = 1}^N k_{ij}\log p_{ij} + (1 - k_{ij})\log (1 - p_{ij})
\end{equation}
where $S$ is the number of reasoning steps of the geometry problem, $k_{ij}$ and $p_{ij}$ are the label/probability for the $j$-th knowledge in $i$-th reasoning step, respectively. The entire loss is the sum of $\mathcal{L}_g$ and $\mathcal{L}_c$: $\mathcal{L} = \mathcal{L}_g + \mathcal{L}_c$.

\section{Experiments}
\begin{table*}[t]
    \centering
    \begin{tabular}{l|ccccc|ccccc}
        \toprule
         & \multicolumn{5}{c}{GeoQA} & \multicolumn{5}{|c}{GeoQA+} \\
        \midrule\midrule
        Methods & Total & Angle & Length & Other & No Result & Total & Angle & Length & Area & No Result \\
        \midrule
        Human Text Only & 63.0 & 58.0 & 71.7 & 55.6 & - & - & - & - & - & - \\
        Human Text-Diagram & 92.3 & 94.2 & 90.5 & 87.0 & - & - & - & - & - & - \\
        \midrule
        CogVLM-17B & 7.96 & 8.39 & 7.77 & 5.55 & 65.2 & 8.09 & 7.45 & 10.1 & 6.15 & 63.0 \\
        VisualGLM-6B & 10.6 & 9.60 & 12.4 & 9.26 & 62.5 & 11.2 & 10.6 & 0.00 & 25.0 & 62.0 \\ 
        ChatGPT & 12.1 & 13.9 & 9.54 & 11.1 & 61.4 & 18.0 & 18.1 & 17.7 & 18.5 & 60.1 \\
        GPT-4 & 26.4 & 25.9 & 29.3 & 14.8 & 56.6 & 30.4 & 23.9 & 35.9 & 38.5 & 51.2 \\
        GPT-4V & 29.7 & 28.1 & 32.5 & 27.8 & 51.4 & 28.5 & 24.2 & 29.0 & 40.0 & 52.5 \\
        GPT-4 + Knowledge & 32.2 & 33.1 & 32.2 & 25.9 & 50.9 & 32.4 & 22.9 & 42.7 & 40.0 & 49.5 \\
        GPT-4V + Knowledge & 34.4 & 32.9 & 35.7 & 38.9 & 49.2 & 31.6 & 25.5 & 37.5 & 37.7 & 48.7 \\
        \midrule
        Seq2Prog & 60.7 & 71.2 & 49.1 & 40.7 & 17.0 & 57.2 & 57.7 & 52.0 & 65.4 & 20.3 \\
        BERT2Prog & 57.6 & 67.9 & 45.6 & 40.7 & 14.6 & 56.6 & 57.4 & 54.4 & 58.5 & 19.0 \\
        RoBERTa2Prog & 60.7 & 71.7 & 48.0 & 42.6 & 16.7 & 59.4 & 59.3 & 58.5 & 61.5 & 19.0 \\
        \midrule
        NGS$^\#$ & 61.9 & 72.4 & 49.8 & 40.7 & 13.7 & 59.1 & 59.0 & \underline{59.7} & 56.9 & 16.8 \\
        Geoformer$^\#$ & 62.7 & \underline{74.1} & 49.8 & 40.7 & 16.3 & 60.2 & 60.9 & 58.1 & 62.3 & 17.2 \\
        DPE-NGS$^\#$ & 63.4 & 72.7 & \underline{53.0} & 44.4 & \underline{12.7} & 60.9 & 62.5 & 54.0 & \textbf{66.9} & \underline{15.2} \\
        SCA-GPS$^\#$ & \underline{63.7} & \textbf{75.0} & 49.8 & \underline{46.8} & 13.0 & \underline{61.8} & \underline{62.6} & 58.9 & 64.6 & 15.9 \\
        \textbf{DualGeoSolver} & \textbf{65.2}$^*$ & 73.6 & \textbf{55.1} & \textbf{48.2} & \textbf{12.2} & \textbf{65.1}$^*$ & \textbf{65.2} & \textbf{63.7} & \underline{66.2} & \textbf{14.6} \\
        \bottomrule
    \end{tabular}
    \caption{Experimental results of all methods. Methods with $\#$ denote the results are re-produced with the authors’ code.}
    \label{tab:main-results}
\end{table*}

\subsection{Experimental Setup}
\textbf{Datasets.} We conduct experiments on two public available datasets: GeoQA and GeoQA+.
The problems in GeoQA are classified into three categories: Angle, Length, and Other, and problems in GeoQA+ are also classified into three categories: Angle, Length and Area. GeoQA+ dataset is an extension of GeoQA dataset, which includes additional problems with a higher average number of reasoning steps and a wider range of problem types, making them more challenging to solve \cite{Cao2022AnAB}.

Since Knowledge Selection Module retrieves knowledge for each reasoning step, it requires knowledge annotations on each individual reasoning step for training. However, both datasets only provide annotations for the entire problem without specifying them to each individual reasoning step. Fortunately, both datasets provide a detailed textual analysis for each problem, which allows us to leverage ChatGPT \footnote{\href{https://chat.openai.com/}{https://chat.openai.com/}} to assign a subset of knowledge to each reasoning step.


\textbf{Implementation Details.}
During training, we keep the parameter of diagram encoder unchanged, and we set the learning rate of RoBERTa to $2e^{-5}$, the learning rate of multi-modal fusion module and Goal Generation Module (GGM) to $1e^{-5}$, and the learning rate of other modules to $1e^{-3}$. We use Adam as the optimizer and set the batch size as 32 while training. The total training epochs is set to 100. All experiments were conducted on an NVIDIA A6000 GPU, with PyTorch version 1.13.1.

While testing, we apply beam-search strategy with beam size $B = 10$ to generate program sequences and utilize a program executor to execute these program sequences in descending order of their predicted probabilities to obtain numerical results. It selects the program sequence that is first successfully executed and matches an option $c_i$ in the candidates $\boldsymbol{c} = \{c_1, c_2, c_3, c_4\}$ as the final solution. If all $B$ program sequences fail or none of them matches an option in the candidates, the executor will report ``No Result" instead of guessing an option.

\textbf{Baselines.} We compare our DualGeoSolver with baselines from 3 classes, including: general Seq2Seq methods, GPS specified neural solvers and large language models. Specifically, general Seq2Seq methods include a LSTM-based model (LSTM2Prog) \cite{Hochreiter1997LongSM}, a BERT-based model (BERT2Prog) \cite{devlin2018bert} and a RoBERTa-based model (RoBERTa2Prog) \cite{Liu2019RoBERTaAR}. GPS specified solvers are NGS \cite{Chen2021GeoQAAG}, Geoformer \cite{Chen2022UniGeoUG}, DPE-NGS \cite{Cao2022AnAB} and SCA-GPS \cite{Ning2023ASC}. LLMs include methods utilize textual inputs only, including ChatGPT and GPT-4 \cite{achiam2023gpt}, and methods utilize both text and diagram, including CogVLM-17B \cite{wang2023cogvlm}, VisualGLM-6B \cite{ding2021cogview,du2022glm} and GPT-4V. We also provide LLMs with geometry knowledge extracted from $K_G$, which are denoted as ``GPT-4(V) + Knowledge''.
The prompt for instructing LLMs to solve geometry problems is designed in a one-shot approach:

\definecolor{textblock_bg}{RGB}{240,255,255}
\begin{tcolorbox}[width=\linewidth, colback=textblock_bg, colframe=white, left=8pt,right=8pt,top=4pt,bottom=4pt]
\texttt{Please solve the geometry problem based on the following problem text. After solving the problem, provide the final calculation result individually at the end, and the final result should be a numerical number or a LaTeX expression, for example: "[Final Result]: 120".}

\texttt{Here is an example of solving geometry problems: \{example\}.}

\texttt{Problem: \{problem\}.}
\end{tcolorbox}

For GPT-4(V) + Knowledge, we additionally provide the ground truth geometry knowledge to the example and the problem to solve. The geometry knowledge along with their explanations are obtained from $K_G$. The prompt comes to be:

\begin{tcolorbox}[width=\linewidth, colback=textblock_bg, colframe=white, left=8pt,right=8pt,top=4pt,bottom=4pt]
\texttt{\{Instruction and Example\} (Same as the prompt above).}

\texttt{Problem: \{problem\}.}

\texttt{Relevant knowledge explanations: \{geometry knowledge\}.}
\end{tcolorbox}

Moreover, we also provide the performances of humans in GPS, which are excerpted from \cite{Chen2021GeoQAAG}.

\subsection{Experimental Results}
We show the Total Accuracy (``Total''), accuracy on specific problem categories (e.g., ``Angle''), and No Result Rate (``No Result'') in Table \ref{tab:main-results}\footnote{Due to the different version of PyTorch, we rerun all the baselines with PyTorch version 1.13.1 for a fair comparison.}. First, our DualGeoSolver surpasses all the baselines. By applying paired t-test, its improvements over the SOTA method SCA-GPS on ``Total" of both datasets are statistically significant with $p < 0.01$ (marked with $*$). It demonstrates the rationality and effectiveness of our DualGeoSolver in modeling human reasoning process for GPS. Second, our DualGeoSolver achieves greater performance improvements on the more difficult dataset GeoQA+. It reflects that through modeling human application of geometry knowledge and the goal-oriented reasoning manner in dual-reasoner, our DualGeoSolver has better generalization ability and robustness.
Third, all the GPS specific solvers, except NGS on GeoQA+, outperforms the general Seq2Seq methods. It indicates the importance of diagram in solving geometry problems, which further shows the necessity of our Visual Spotlight Module to capture the relationship between reasoning goals and diagram.
Fourth, LLMs do not perform well in GPS, which we believe is caused by the disparity between geometry diagrams and images used for pre-training LLMs. But we notice that by applying geometry knowledge, GPT-4(V)'s performances have drastically improved, which demonstrates the necessity of introducing knowledge into current methods.

\subsection{Ablation Study}
\begin{table}[t]
    \centering
    \begin{tabular}{l|cc|cc}
        \toprule
         & \multicolumn{2}{|c|}{GeoQA} & \multicolumn{2}{|c}{GeoQA+} \\
        \midrule\midrule
        Models & Total & No Result & Total & No Result \\
        \midrule
        DualGeoSolver & \textbf{65.2} & 12.2 & \textbf{65.1} & \textbf{14.6} \\
        w/o KSM & 63.3 & 14.4 & 63.1 & 15.4 \\
        w/o VSM & 64.4 & \textbf{11.3} & 64.7 & 16.5 \\
        w/o KIM & 64.8 & 13.2 & 64.0 & 15.2 \\
        w/o GGM & 63.3 & 15.2 & 63.6 & 15.1 \\
        \bottomrule
    \end{tabular}
    \caption{Experimental results of ablation study.}
    \label{tab:ablation}
\end{table}

To verify the effectiveness of modules in our DualGeoSolver, we conduct the ablation study in Table \ref{tab:ablation}. Specifically, for Knowledge System, ``w/o KSM'' omits Knowledge Selection Module, thus ignores the geometry knowledge application in reasoning and the loss $\mathcal{L}_c$ in Eq.~\eqref{loss_c} in training. ``w/o VSM'' ignores Visual Spotlight Module that captures visual information related to reasoning goal and only inputs $H^{K}_t$ into KIM for generating guiding information. ``w/o KIM'' ignores Knowledge Injection Module, and directly feeds the geometry knowledge $H^{K}_t$, which will be aggregated by $s_{t - 1}$, and visual context vector $h^{vis}_t$ into Token Prediction Module. For Inference System, ``w/o GGM'' omits Goal Generation Module and directly regards $s_t$ as the reasoning goal.

From Table \ref{tab:ablation}, we first notice that the accuracy degrades when any module is missing, which verifies the rationality and necessity of all components in DualGeoSolver. Second, the accuracy degrades the most when KSM is missing, showing that KSM is the most crucial module in DualGeoSolver and applying geometry knowledge in solving geometry problems is important. Third, the performance of DualGeoSolver is significantly hindered without Goal Generation Module, which indicates that it is necessary to develop a human-like goal-oriented mechanism that fully integrates the known conditions and all preceding reasoning information to derive the next reasoning goal. Last but not least, the importance of KIM and VSM varies with the difficulty of the geometry problems. On GeoQA+, which is more challenging than GeoQA, KIM becomes more crucial than VSM. We believe this is because, for harder geometry problems, simply capturing the relationships between geometry primitives in the diagram is not sufficient to provide enough information for solving the problem, where KIM is needed to effectively integrate the diagram information with geometry knowledge.


\subsection{Knowledge Analysis}
\begin{figure}[t]
    \centering
    \includegraphics[width=.5\linewidth]{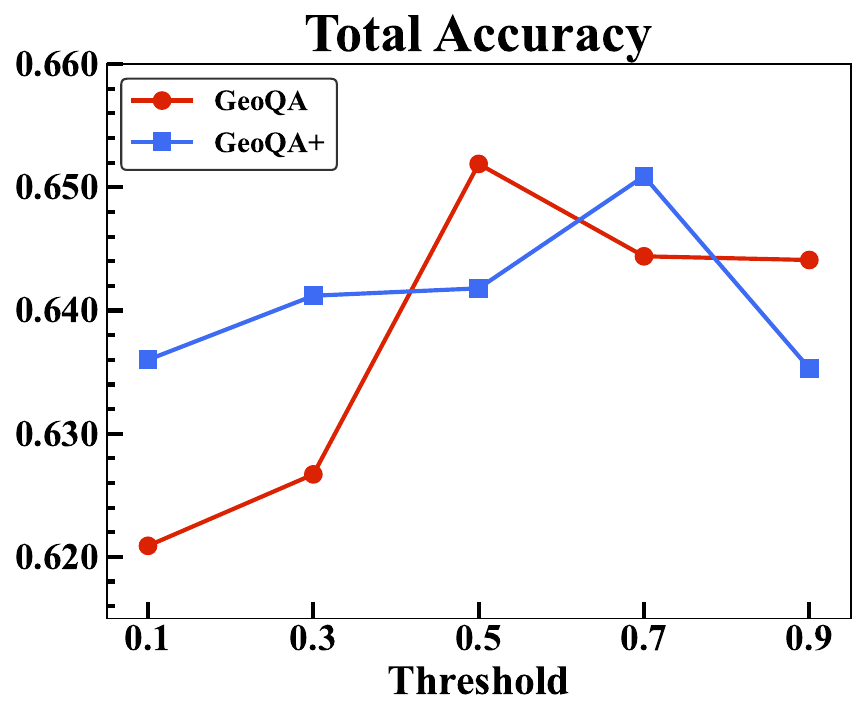}\hfill
    \includegraphics[width=.5\linewidth]{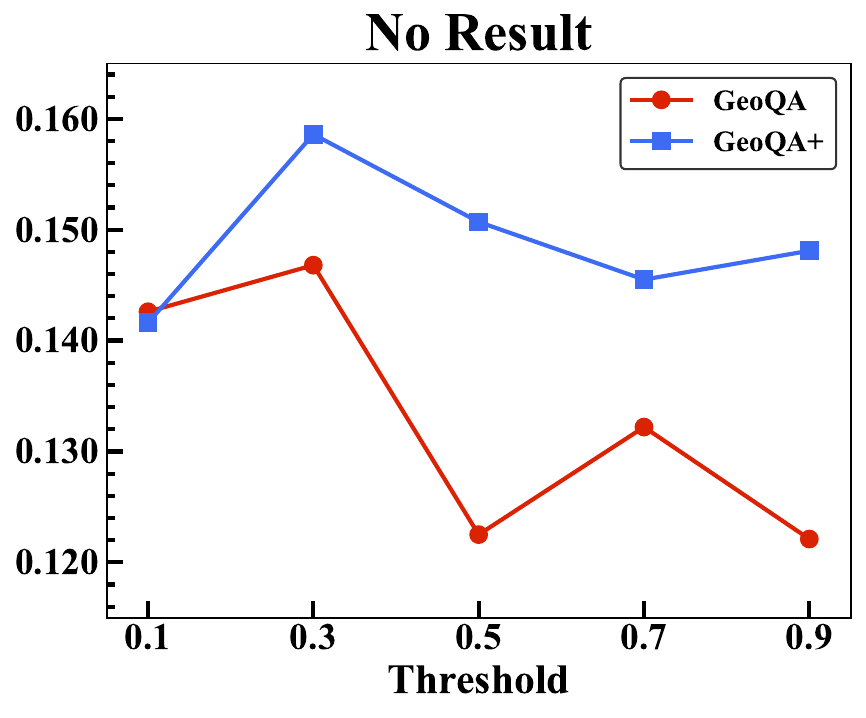}
    \caption{Performance of DualGeoSolver when varying threshold $\theta$.}
    \label{fig:knowledge-threshold}
\end{figure}

\begin{table}
    \centering
    \begin{tabular}{l|cc|cc}
        \toprule
         & \multicolumn{2}{|c|}{GeoQA} & \multicolumn{2}{|c}{GeoQA+} \\
        \midrule\midrule
        Models & OP & Absolute & OP & Absolute \\
        \midrule
        w/o KSM & 56.1 & 44.0 & 44.7 & 31.8 \\
        DualGeoSolver & \textbf{58.4} & \textbf{46.9} & \textbf{45.6} & \textbf{32.4} \\
        \bottomrule
    \end{tabular}
    \caption{Single-step accuracy w/ and w/o geometry knowledge.}
    \label{tab:knowledge-steps}
\end{table}

To verify the role of geometry knowledge in solving geometry problems, we conduct two experiments: evaluating the performance of our DualGeoSolver with varying threshold $\theta$ in Knowledge Selection Module, and assessing the single-step accuracy with/without application of geometry knowledge.

From Figure \ref{fig:knowledge-threshold}, we observe that the performance is better when the threshold $\theta$ is closer to the middle value and decreases when the threshold is too low or too high. This demonstrates that applying excessive incorrect knowledge or missing essential knowledge during reasoning is detrimental to geometry problem solving. Besides, the optimal knowledge threshold for our DualGeoSolver to achieve the best performance varies across different datasets, being $\theta = 0.5$ on GeoQA and $\theta = 0.7$ on GeoQA+. It indicates that our DualGeoSolver prefers the usage of more precise and accurate geometry knowledge on GeoQA+, while tends to use a greater amount of geometry knowledge on GeoQA.

We also assess the single-step accuracy with and without applying geometry knowledge in Table \ref{tab:knowledge-steps}. Specifically, ``OP'' means the accuracy of operator in each reasoning step, and ``Absolute'' means the accuracy of the whole step. For example, for Step 1 in Figure~\ref{fig:intro-case} that requires to reason ``[Minus, C\_3, N\_0]'', if the model correctly derives operator ``Minus'', then ``OP'' will be considered correct; only if all three tokens are correctly generated, ``Absolute'' will be considered correct. From Table \ref{tab:knowledge-steps}, the performances are worse without Knowledge Selection Module on both datasets, which directly demonstrates the effectiveness of applying geometry knowledge in GPS. It is worth noting that the lower accuracy on ``OP'' and ``Absolute'' compared to ``Total Accuracy'' may be due to that model predicts a program that differs from the gold program but can still solve the geometry problem.

\subsection{Case Study}
\begin{figure}[t]
    \centering
    \includegraphics[width=\linewidth]{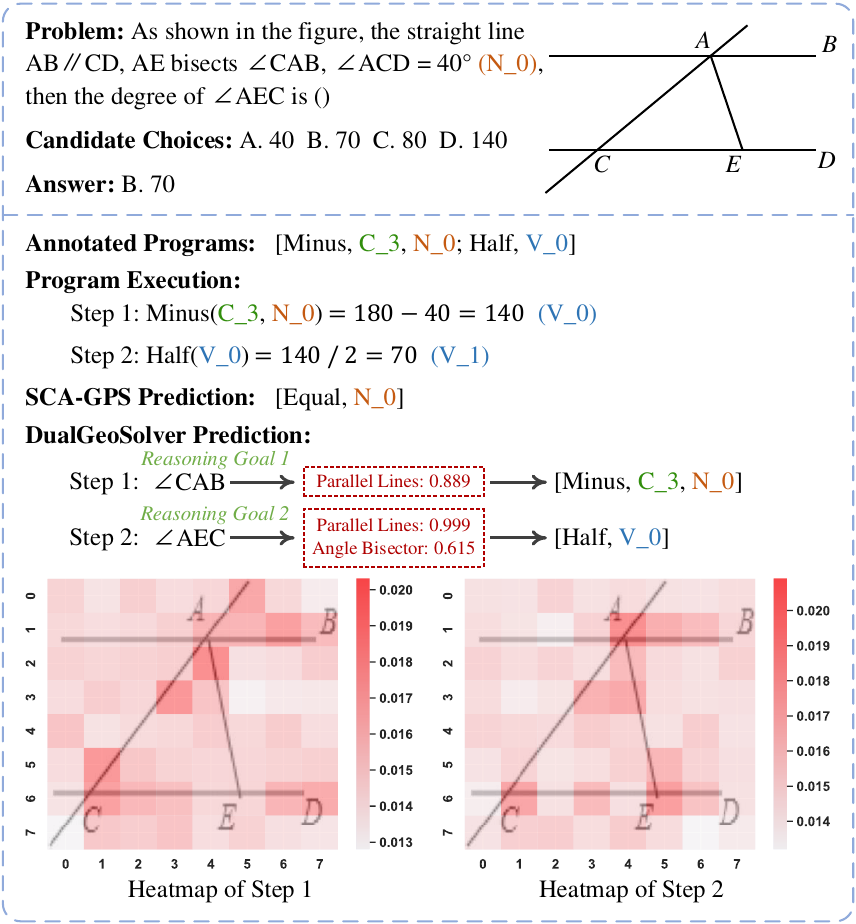}
    \caption{A typical geometry problem in GeoQA dataset, with program sequences generated by different solvers.}
    \label{fig:case-study}
\end{figure}

Further, we present a typical case in Figure \ref{fig:case-study} to verify the interpretability of DualGeoSolver. We first present the problem and program sequences generated by different solvers. Then, we report the knowledge selection probability in knowledge Selection Module and visualize the attention weights of Visual Spotlight Module in our DualGeoSolver. 

This problem requires solvers to understand the properties of parallel lines and be aware that an angle bisector $AE$ divides $\angle{CAB}$ into two equal angles. The program sequence generated by SOTA method SCA-GPS is incorrect, which appears to calculate the corresponding angle of $\angle{ACE}$. Comparatively, our DualGeoSolver achieves a two-step reasoning process. In step 1, its reasoning goal focuses on $\angle{ACE}$, $\angle{CAB}$ and parallel lines $AB$, $CD$ according to the heatmap. Meanwhile, it selects geometry knowledge ``Parallel Lines" with the confidence of $0.889$. Combining them, it correctly deduces the program tokens ``[Minus, C\_3, N\_0]". In step 2, its reasoning goal turns to the angle bisector $AE$ and $\angle{CAB}$, and selects two knowledge ``Parallel Lines" and ``Angle Bisector" simultaneously, and finally deduces the correct program tokens ``[Half, V\_0]". These observations verify the rationality and effectiveness of DualGeoSolver in simulating human reasoning process, which benefits from Knowledge System and Goal Generation Module in dual-reasoner.


\section{Conclusion}
In this paper, we proposed a novel \textbf{Dual}-reasoning \textbf{Geo}metry \textbf{Solver} \textbf{(DualGeoSolver)}, which drew insights from dual-process theory and built Knowledge-Inference systems to conduct human-like dual-reasoning. Knowledge System managed geometry knowledge and diagram information that aided reasoning. Inference System adopted a goal-oriented reasoning mechanism and applied knowledge to program generation. Experiments on GeoQA and GeoQA+ datasets demonstrated the advantages of DualGeoSolver in answer accuracy and robustness.
For future studies, we will enhance the ability of LLMs by introducing reasoning goals and geometry knowledge to achieve more reliable and interpretable GPS.

\section*{Acknowledgement}
This research was partially supported by grants from the National Natural Science Foundation of China (Grants No. 62106244, No. U20A20229), and the University Synergy Innovation Program of Anhui Province (GXXT-2022-042).

\bibliographystyle{named}
\bibliography{ijcai24}

\end{document}